\documentclass[11pt,a4paper]{article}
\usepackage[hyperref]{acl2020}
\usepackage{times}
\usepackage{latexsym}

\usepackage{microtype}
\usepackage{enumitem}
\usepackage{graphicx}
\usepackage{booktabs}  
\usepackage{textcomp}
\usepackage{threeparttable}
\usepackage{adjustbox}
\usepackage{caption}    
\usepackage{subcaption} 

\aclfinalcopy 

\makeatletter
\def\blfootnotestar{\gdef\@thefnmark{*}\@footnotetext}
\makeatother

\makeatletter
\def\blfootnotedagger{\gdef\@thefnmark{$\dagger$}\@footnotetext}
\makeatother

\title{BEEP! Korean Corpus of Online News Comments\\for Toxic Speech Detection}

\author{Jihyung Moon\textsuperscript{*,$\dagger$,1}, Won Ik Cho\textsuperscript{*,2}, Junbum Lee\textsuperscript{3} \\
  Department of Industrial Engineering\textsuperscript{1}, \\
  Department of Electrical and Computer Engineering and INMC\textsuperscript{2}, \\
  Graduate School of Data Science\textsuperscript{3}, \\Seoul National University, Seoul \\
   \texttt{\{ans1107,tsatsuki,beomi\}@snu.ac.kr}\\
}

\date{}

\begin{document}
\maketitle
\blfootnotestar{Both authors contributed equally to this manuscript.}
\blfootnotedagger{This work was done after the graduation.}

\begin{abstract}
Toxic comments in online platforms are an unavoidable social issue under the cloak of anonymity. Hate speech detection has been actively done for languages such as English, German, or Italian, where manually labeled corpus has been released. In this work, we first present 9.4K manually labeled entertainment news comments for identifying Korean toxic speech, collected from a widely used online news platform in Korea.
The comments are annotated regarding social bias and hate speech since both aspects are correlated. The inter-annotator agreement Krippendorff’s alpha score is 0.492 and 0.496, respectively. We provide benchmarks using CharCNN, BiLSTM, and BERT, where BERT achieves the highest score on all tasks. The models generally display better performance on bias identification, since the hate speech detection is a more subjective issue. 
Additionally, when BERT is trained with bias label for hate speech detection, the prediction score increases, implying that bias and hate are intertwined.
We make our dataset publicly available and open competitions with the corpus and benchmarks.
\end{abstract}

\section{Introduction}
Online anonymity provides freedom of speech to many people and lets them speak their opinions in public. However, anonymous speech also has a negative impact on society and individuals~\citep{banks2010regulating}. 
With anonymity safeguards, individuals easily express hatred against others based on their superficial characteristics such as gender, sexual orientation, and age~\citep{elsherief2018hate}. Sometimes the hostility leaks to the well-known people who are considered to be the representatives of targeted attributes.

Recently, Korea had suffered a series of tragic incidents of two young celebrities that are presumed to be caused by toxic comments~\citep{fortin2019,McCurry2019b,McCurry2019}. Since the incidents, two major web portals in Korea decided to close the comment system in their entertainment news aggregating service~\citep{Yeo2019,Yim2020}. Even though the toxic comments are now avoidable in those platforms, the fundamental problem has not been solved yet.

To cope with the social issue, we propose the first Korean corpus annotated for toxic speech detection. Specifically, our dataset consists of 9.4K comments from Korean online entertainment news articles. 
Each comment is annotated on two aspects, the existence of social bias and hate speech, given that hate speech is closely related to bias~\citep{boeckmann2002understanding,waseem2016hateful,davidson2017automated}.
Considering the context of Korean entertainment news where public figures encounter stereotypes mostly intertwined with gender, we weigh more on the prevalent bias. For hate speech, our label  categorization refers that of \citet{davidson2017automated}, namely \textit{hate}, \textit{offensive}, and \textit{none}.

The main contributions of this work are as follows:
\begin{itemize}[noitemsep]
    \item We release the first Korean corpus manually annotated on two major toxic attributes, namely bias and hate\footnote{https://github.com/kocohub/korean-hate-speech}.
    \smallskip 
    \item We hold Kaggle competitions\footnote{www.kaggle.com/c/korean-gender-bias-detection}\footnote{www.kaggle.com/c/korean-bias-detection}\footnote{www.kaggle.com/c/korean-hate-speech-detection} and provide benchmarks to boost further research development. 
    \smallskip
    \item We observe that in our study, hate speech detection benefits the additional bias context.
\end{itemize}

\section{Related Work}
\label{section2-related-work}
The construction of hate speech corpus has been explored for a limited number of languages,  
such as English~\citep{waseem2016hateful,davidson2017automated,zampieri2019semeval,basile2019semeval}, Spanish~\citep{basile2019semeval}, Polish~\citep{ptaszynski2019results}, Portuguese~\citep{fortuna2019hierarchically}, and Italian~\citep{sanguinetti2018italian}. 

For Korean, works on abusive language have mainly focused on the qualitative discussion of the terminology~\citep{hong2016research}, whereas reliable and manual annotation of the corpus has not yet been undertaken.
Though profanity termbases are currently available\footnote{https://github.com/doublems/korean-bad-words}\footnote{https://github.com/LDNOOBW/List-of-Dirty-Naughty-Obscene-and-Otherwise-Bad-Words}, term matching approach frequently makes false predictions (e.g., neologism, polysemy, use-mention distinction), and more importantly, not all hate speech are detectable using such terms~\citep{zhang2018detecting}.

In addition, hate speech is situated within the context of social bias~\citep{boeckmann2002understanding}. 
\citet{waseem2016hateful} and \citet{davidson2017automated} attended to bias in terms of hate speech, however, their interest was mainly in texts that explicitly exhibit sexist or racist terms.
In this paper, we consider both explicit and implicit stereotypes, and scrutinize how these are related to hate speech.

\section{Collection}

We constructed the Korean hate speech corpus using the comments from a popular domestic entertainment news aggregation platform. Users had been able to leave comments on each article before the recent overhaul~\citep{Yim2020}, and we had scrapped the comments from the most-viewed articles.

In total, we retrieved 10,403,368 comments from 23,700 articles published from January 1, 2018 to February 29, 2020. We draw 1,580 articles using stratified sampling and extract the top 20 comments ranked in the order of Wilson score~\citep{wilson1927probable} on the downvote for each article. Then, we remove duplicate comments, single token comments (to eliminate ambiguous ones), and comments composed with more than 100 characters (that could convey various opinions). Finally, 10K comments are randomly selected among the rest for annotation. 

We prepared other 2M comments by gathering the top 100 sorted with the same score for all articles and removed with any overlaps regarding the above 10K comments. This additional corpus is distributed without labels, expected to be useful for pre-training language models on Korean online text.

\begin{figure}
	\centering
	\includegraphics[width=\columnwidth]{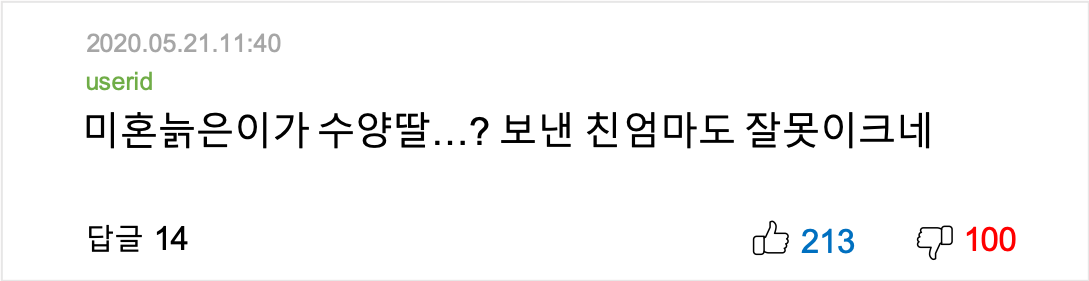}
	\caption{A sample comment from the online news platform. It is composed of six parts: written date and time, masked user id, content, the number of replies, and the number of up/down votes (from top left to bottom right).} 
	\label{fig:naver-news-comments-example}
\end{figure}

\section{Annotation}

The annotation was performed by 32 annotators consisting of 29 workers from a crowdsourcing platform \textit{DeepNatural AI}\footnote{https://app.deepnatural.ai/} and three natural language processing (NLP) researchers. Every comment was provided to three random annotators to assign the majority decision. Annotators are asked to answer two three-choice questions for each comment:

\begin{enumerate}[noitemsep]
    \item What kind of bias does the comment contain?
    \begin{itemize}
        \item \textit{Gender bias, Other biases}, or \textit{None}
    \end{itemize}
    \item Which is the adequate 
    category for the comment in terms of hate speech?
    \begin{itemize}
        \item \textit{Hate, Offensive}, or \textit{None}
    \end{itemize}
\end{enumerate}

They are allowed to skip comments which are too ambiguous to decide. Detailed instructions are described in Appendix~\ref{appendix:guideline}. 
Note that this is the first guideline of social bias and hate speech on Korean online comments.

\subsection{Social Bias}

Since hate speech is situated within the context of social bias~\citep{boeckmann2002understanding}, we first identify the bias implicated in the comment. Social bias is defined as a preconceived evaluation or prejudice towards a person/group with certain social characteristics: gender, political affiliation, religion, beauty, age, disability, race, or others. Although our main interest is on gender bias, other issues are not to be underestimated. Thus, we separate bias labels into three: whether the given text contains gender-related bias, other biases, or none of them. Additionally, we introduce a binary version of the corpus, which counts only the gender bias, that is prevalent among the entertainment news comments.

The inter-annotator agreement (IAA) of the label is calculated based on Krippendorff's alpha~\citep{krippendorff2011computing} that takes into account an arbitrary number of annotators labeling any number of instances. IAA for the ternary classes is 0.492, which means that the agreement is moderate. For the binary case, we obtained 0.767, which implies that the identification of gender and sexuality-related bias reaches quite a substantial agreement.

\subsection{Hate Speech}

Hate speech is difficult to be identified, especially for the comments which are context-sensitive. Since annotators are not given additional information, labeling would be diversified due to the difference in pragmatic intuition and background knowledge thereof. To collect reliable hate speech annotation, we attempt to establish a precise and clear guideline. 

We consider three categories for hate speech: \textit{hate}, \textit{offensive but not hate}, and \textit{none}. As socially agreed definition lacks for Korean\footnote{Though a government report is available for the Korean language~\citep{hong2016research}, we could not reach a fine extension to the quantitative study on online spaces.}, we refer to the hate speech policies of  \citet{youtube2020hatespeech,facebook2020hatespeech,twitter2020hatespeech}. Drawing upon those, we define hate speech in our study as follows:
\begin{itemize}[noitemsep]
    \item If a comment explicitly expresses hatred against individual/group based on any of the following attributes: sex, gender, sexual orientation, gender identity, age, appearance, social status, religious affiliation, military service, disease or disability, ethnicity, and national origin
    \item If a comment severely insults or attacks individual/group; this includes sexual harassment, humiliation, and derogation
\end{itemize} 
However, note that not all the rude or aggressive comments necessarily belong to the above definition, as argued in \citet{davidson2017automated}. We often see comments that are offensive to certain individuals/groups in a qualitatively different manner. We identify these as offensive and set the boundary as follows:
\begin{itemize}[noitemsep]
    \item If a comment conveys sarcasm via rhetorical expression or irony
    \item If a comment states an opinion in an unethical, rude, coarse, or uncivilized manner
    \item If a comment implicitly attacks individual/group while leaving rooms to be considered as freedom of speech
\end{itemize}

The instances that do not meet the boundaries above were categorized as \textit{none}. The IAA on the hate categories is $\alpha$ = 0.496, which implies a moderate agreement.

\section{Corpus}

\begin{table}[]
\centering
\resizebox{0.9\columnwidth}{!}{%
\begin{tabular}{@{}c|ccc|c@{}}
\toprule
\textbf{(\%)} & \textbf{Hate} & \textbf{Offensive} & \textbf{None} & \textbf{Sum (Bias)} \\ \midrule
\textbf{Gender}       & 10.15         & 4.58               & 0.98          & 15.71               \\ 
\textbf{Others}        & 7.48          & 8.94               & 1.74          & 18.16               \\ 
\textbf{None}         & 7.48          & 19.13              & 39.08         & 65.70               \\ \midrule
\textbf{Sum (Hate)}   & 25.11         & 32.66              & 41.80         & 100.00               \\ \bottomrule
\end{tabular}%
}
\caption{Distribution of the annotated corpus.}
\label{tab:corpus-distribution}
\end{table}

\paragraph{Release} 
From the 10k manually annotated corpus, we discard 659 instances that are either skipped or failed to reach an agreement. 
We split the final dataset into the train (7,896), validation (471), and test set (974) and released it on the Kaggle platform to leverage the leaderboard system. For a fair competition, labels on the test set are not disclosed. Titles of source articles for each comment are also provided, to help participants exploit context information. 

\paragraph{Class distribution} 
Table~\ref{tab:corpus-distribution} depicts how the classes are composed of. The bias category distribution in our corpus is skewed towards \textit{none}, while that of \textit{hate} category is quite balanced. 
We also confirm that the existence of hate speech is correlated with the existence of social bias. In other words, when a comment incorporates a social bias, it is likely to contain hate or offensive speech.

\begin{figure*}[t!]
\centering
\captionsetup[subfigure]{oneside,margin={0.62cm,0cm}}
    \begin{subfigure}[b]{0.75\columnwidth}
        \centering
    \includegraphics[width=0.9\textwidth]{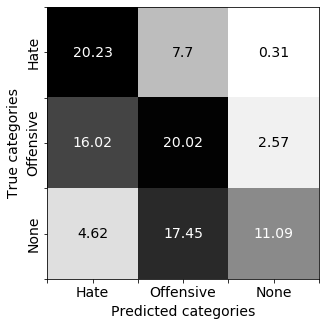}
        \caption{BERT predictions}
    \end{subfigure}
    \begin{subfigure}[b]{0.75\columnwidth}
        \centering
    \includegraphics[width=0.9\textwidth]{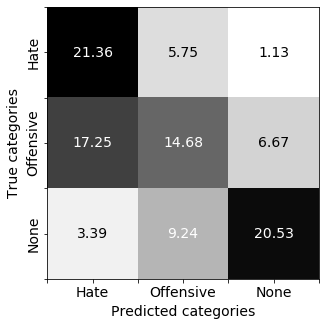}
        \caption{BERT predictions with bias label}
    \end{subfigure}
\caption{Confusion matrix on the model inference of hate categories.}
\label{fig:confusion-matrix}
\end{figure*}

\section{Benchmark Experiment}

\subsection{Models}

We implemented three baseline classifiers: character-level convolutional neural network (CharCNN)~\citep{zhang2015character}, bidirectional long short-term memory (BiLSTM)~\citep{schuster1997bidirectional}, and bidirectional encoder representations from Transformer (BERT)~\citep{devlin2018bert} based model.
For BERT, we adopt KoBERT\footnote{https://github.com/SKTBrain/KoBERT}, a pre-trained module for the Korean language, and apply its tokenizer to BiLSTM as well. 
The detailed configurations are provided in Appendix~\ref{appendix:config}, and we additionally report the term matching approach using the aforementioned profanity terms to compare with the benchmarks.

\subsection{Results}

\begin{table}[]
\centering
\resizebox{0.9\columnwidth}{!}{%
\begin{tabular}{@{}c|ccc@{}}
\toprule
\textbf{F1}               & \textbf{Bias (binary)} & \textbf{Bias (ternary)} & \textbf{Hate} \\ \midrule
Term Matching   &        -            &       -         &   0.195              \\ \midrule
CharCNN & 0.547              & 0.535               & 0.415       \\
BiLSTM  & 0.302                      & 0.291                        & 0.340       \\
BERT    & \textbf{0.681}              & \textbf{0.633}               & \textbf{0.525}       \\ \midrule
BERT (+ bias) & - & - & \textbf{0.569} \\
\bottomrule
\end{tabular}%
}
\caption{F1 score of benchmarks on the test set. Note that the term matching model checks the presence of hate or offensiveness. Therefore, in this case, we combine \textit{hate} and \textit{offensive} into a single category, turning the original ternary task into binary.
}
\label{tab:baseline}
\end{table}

Table~\ref{tab:baseline} depicts F1 score of the three baselines and the term matching model. 
The results demonstrate that the models trained on our corpus have an advantage over the term matching method.
Compared with the benchmarks, BERT achieves the best performance for all the three tasks: binary and ternary bias identification tasks, and hate speech detection. 
Each model not only shows different performances but also presents different characteristics.

\paragraph{Bias detection}
When it comes to the gender-bias detection, the task benefits more on CharCNN than BiLSTM since the bias label is highly correlated with frequent gender terms (e.g., \textit{he, she, man, woman, ...}) in the dataset. It is known that CharCNN well captures the lexical components that are present in the document.

However, owing to that nature, CharCNN sometimes yields results that are overly influenced by the specific terms which cause false predictions.
For example, the model fails to detect bias in \textit{``What a long life for a GAY"} but guesses \textit{``I think she is the prettiest among all the celebs''} to contain bias. CharCNN overlooks \textit{GAY} while giving a wrong clue due to the existence of female pronouns, namely \textit{she} in the latter.

Similar to the binary prediction task, CharCNN outperforms BiLSTM on ternary classification. Table~\ref{tab:bias-ternary} demonstrates that BiLSTM hardly identifies \textit{gender} and \textit{other} biases.

\begin{table}[]
\centering
\resizebox{0.9\columnwidth}{!}{%
\begin{tabular}{@{}c|ccc|c@{}}
\toprule
\textbf{F1}               & \textbf{Gender} & \textbf{Others} & \textbf{None} & \textbf{Bias (ternary)} \\ \midrule
CharCNN & 0.519             & 0.259            & 0.826  &   0.535       \\
BiLSTM  & 0.055             & 0.000            & 0.819  &   0.291       \\
BERT    & 0.693             & 0.326            & 0.880  & \textbf{0.633} \\
\bottomrule
\end{tabular}%
}
\caption{Detailed results on macro-F1 of Bias (ternary)}
\label{tab:bias-ternary}
\end{table}

BERT detects both biases better than the other models. From the highest score obtained by BERT, we found that rich linguistic knowledge and semantic information is helpful for bias recognition. 

We also observed that all the three models barely perform well on \textit{others} (Table~\ref{tab:bias-ternary}). 
To make up a system that covers the broad definition of \textit{other} bias, it would be better to predict the label as the non-\textit{gender} bias.
For instance, it can be performed as a two-step prediction: the first step to distinguish whether the comment is biased or not and the second step to determine whether the biased comment is gender-related or not.

\paragraph{Hate speech detection}
For hate speech detection, all models faced performance degradation compared to the bias classification task, since the task is more challenging. Nonetheless, BERT is still the most successful, and we conjecture that hate speech detection also utilizes high-level semantic features. The significant performance gap between term matching and BERT explains how much our approach compensates for the false predictions mentioned in Section~\ref{section2-related-work}. 

Provided \textit{bias} label prepend to each comment as a special token, BERT exhibits better performance. As illustrated in Figure~\ref{fig:confusion-matrix}, additional bias context helps the model to distinguish \textit{offensive} and \textit{none} clearly. This implies our observation on the correlation between bias and hate is empirically supported.

\section{Conclusions}

In this data paper, we provide an annotated corpus that can be practically used for analysis and modeling on Korean toxic language, including hate speech and social bias. In specific, we construct a corpus of a total of 9.4K comments from online entertainment news service. 

Our dataset has been made publicly accessible with baseline models. We launch Kaggle competitions using the corpus, which may facilitate the studies on toxic speech and ameliorate the cyber-bullying issues. 
We hope our initial efforts can be supportive not only to NLP for social good, but also as a useful resource for discerning implicit bias and hate in online languages.

\section*{Acknowledgments}

We greatly thank Hyunjoong Kim for providing financial support and Sangwoong Yoon for giving helpful comments.

\bibliography{acl2020}
\bibliographystyle{acl_natbib}

\appendix

\section{Annotation Guideline}
\label{appendix:guideline}

\subsection{Existence of social bias}

The first property is to note which social bias is implicated in the comment. Here, social bias means hasty guess or prejudice that `a person/group with a certain social identity will display a certain characteristic or act in a biased way'. The three labels of the question are as follows.

\begin{enumerate}[noitemsep]
    \item Is there a gender-related bias, either explicit or implicit, in the text?
    \begin{itemize}
        \item If the text includes bias for gender role, sexual orientation, sexual identity, and any thoughts on gender-related acts (e.g., ``\textit{Wife must be obedient to her husband's words}'', or ``\textit{Homosexual person will be prone to disease.}'')
    \end{itemize}
	\item Are there any other kinds of bias in the text? 
	\begin{itemize}
        \item Other kinds of factors that are considered not gender-related but social bias, including race, background, nationality, ethnic group, political stance, skin color, religion, handicaps, age, appearance, richness, occupations, the absence of military service experience\footnote{Frequently observable in Korea, where the military service is mandatory for males.}, etc.
    \end{itemize}
	\item A comment that does not incorporate the bias
\end{enumerate}

\subsection{Amount of hate, insulting, or offense}

The second property is how aggressive the comment is. 
Since the level of ``aggressiveness'' depends on the linguistic intuition of annotators, we set the following categorization to draw a borderline as precise as possible.

\begin{enumerate}[noitemsep]
    \item Is strong hate or insulting towards the article's target or related figures, writers of the article or comments, etc. displayed in a comment?
    \begin{itemize}
        \item In the case of insulting, it encompasses an expression that can severely harm the social status of the recipient.
        \item In the case of hate, it is defined as an expression that displays aggressive stances towards individuals/groups with certain characteristics (gender role, sexual orientation, sexual identity, any thoughts on gender-related acts, race, background, nationality, ethnic group, political stance, skin color, religion, handicaps, age, appearance, richness, occupations, the absence of military service experience, etc.).
        \item Additionally, it can include sexual harassment, notification of offensive rumors or facts, and coined terms for bad purposes or in bad use, etc.
        \item Just an existence of bad words in the document does not always fall into this category.
    \end{itemize}
	\item Although a comment is not as much hateful or insulting as the above, does it make the target or the reader feel offended?
    \begin{itemize}
        \item It may contain rude or aggressive contents, such as bad words, though not to the extent of hate or insult.
        \item It can emit sarcasm through rhetorical questions or irony.
        \item It may encompass an unethical expression (e.g., jokes or irrelevant questions regarding the figures who passed away).
        \item A comment conveying unidentified rumors can belong to this category.
    \end{itemize}
    \item A comment that does not incorporate any hatred or insulting
\end{enumerate}

\section{Model Configuration}
\label{appendix:config}

Note that each model's configuration is the same for all tasks except for the last layer.

\subsection{CharCNN}

For character-level CNN, no specific tokenization was utilized. The sequence of Hangul characters was fed into the model at a maximum length of 150. The total number of characters was 1,685, including `[UNK]' and `[PAD]' token, and the embedding size was set to 300. 10 kernels were used, each with the size of [3,4,5]. At the final pooling layer, we used a fully connected network (FCN) of size 1,140, with a 0.5 dropout rate~\cite{srivastava2014dropout}. The training was done for 6 epochs.

\subsection{BiLSTM}

For bidirectional LSTM, we had a vocab size of 4,322, with a maximum length of 256. We used BERT SentencePiece tokenizer~\cite{kudo2018sentencepiece}. The width of the hidden layers was 512 (=$256 \times 2$), with four stacked layers. The dropout rate was set to 0.3. An FCN of size 1,024 was appended to the BiLSTM output to yield the final softmax layer. We trained the model for 15 epochs.

\subsection{BERT}

For BERT, a built-in SentencePiece tokenizer of KoBERT was adopted, which was also used for BiLSTM. We set a maximum length at 256 and ran the model for 10 epochs.

\end{document}